\documentclass{ieeeaccess}
\usepackage{cite}
\usepackage{url}
\usepackage{amsmath,amssymb,amsfonts}
\usepackage[ruled,vlined]{algorithm2e}
\usepackage{multirow}
\usepackage{makecell}
\usepackage{algorithmic}
\usepackage{graphicx}
\usepackage{textcomp}
\usepackage{subfigure}
\usepackage[implicit=false]{hyperref}
\usepackage[switch]{lineno}
\def\BibTeX{{\rm B\kern-.05em{\sc i\kern-.025em b}\kern-.08em
    T\kern-.1667em\lower.7ex\hbox{E}\kern-.125emX}}
\begin{document}
\history{Date of publication xxxx 00, 0000, date of current version xxxx 00, 0000.}
\doi{10.1109/ACCESS.2021.3122273}

\title{Improving BERT with Self-Supervised Attention}
\author{\uppercase{Yiren Chen}\href{https://orcid.org/0000-0001-5608-5144}{\includegraphics[scale=0.08]{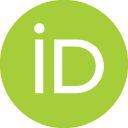}}, 
\uppercase{Xiaoyu Kou}\href{https://orcid.org/0000-0002-1238-967X}{\includegraphics[scale=0.08]{ORCIDiD.png}}, \uppercase{Jiangang Bai}\href{https://orcid.org/0000-0002-2299-2464}{\includegraphics[scale=0.08]{ORCIDiD.png}}, \uppercase{and Yunhai Tong}\href{https://orcid.org/0000-0001-8735-2516}{\includegraphics[scale=0.08]{ORCIDiD.png}}}
\address{Key Laboratory of Machine Perception (MOE), School of EECS, Peking University, Beijing 100871, China}
\tfootnote{This work was supported by the National Key Research and Development Program of China under Grant 2020YFB2103402.}


\markboth
{Author \headeretal: Preparation of Papers for IEEE TRANSACTIONS and JOURNALS}
{Author \headeretal: Preparation of Papers for IEEE TRANSACTIONS and JOURNALS}

\corresp{Corresponding author: Yunhai Tong (e-mail: yhtong@pku.edu.cn).}

\begin{abstract}
One of the most popular paradigms of applying large pre-trained NLP models such as BERT is to fine-tune it on a smaller dataset. However, one challenge remains as the fine-tuned model often overfits on smaller datasets. 
A symptom of this phenomenon is that irrelevant or misleading words in the sentence, which are easy to understand for human beings, can substantially degrade the performance of these fine-tuned BERT models.
In this paper, we propose a novel technique, called Self-Supervised Attention (SSA) to help facilitate this generalization challenge. Specifically, SSA automatically generates weak, token-level attention labels iteratively by probing the fine-tuned model from the previous iteration. We investigate two different ways of integrating SSA into BERT and propose a hybrid approach to combine their benefits. Empirically, through a variety of public datasets, we illustrate significant performance improvement using our SSA-enhanced BERT model.
\end{abstract}


\titlepgskip=-15pt

\maketitle

\section{Introduction}
\label{sec:introduction}
\PARstart{T}{he} models based on self-attention such as Transformer~\cite{vaswani2017attention} have shown their effectiveness for a variety of NLP tasks. One popular use is to leverage a single pre-trained language model, e.g., BERT~\cite{devlin2019bert}, and transfer it to a specific downstream task. However, opportunities remain to further improve these fine-tuned models as many of them often overfit, especially on smaller datasets.

\begin{figure*}[ht]
	\centering
	\subfigure[Original sentence]{
        \begin{minipage}[t]{0.5\linewidth}
        \centering
        \includegraphics[width=0.8\linewidth]{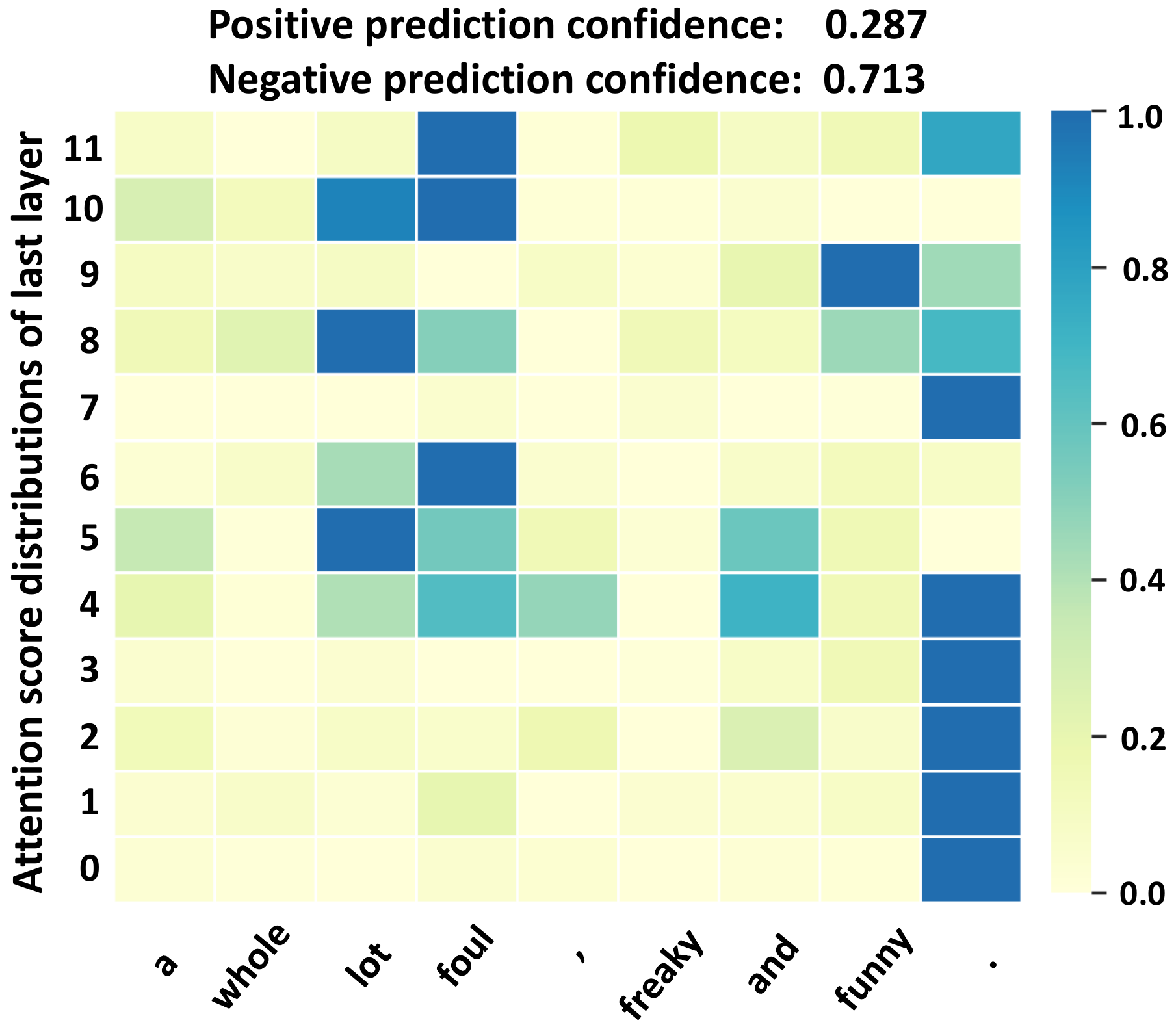}
        \label{fig:original}
    \end{minipage}%
    }%
    \subfigure[Sentence after masking a word]{
        \begin{minipage}[t]{0.5\linewidth}
        \centering
        \includegraphics[width=0.8\linewidth]{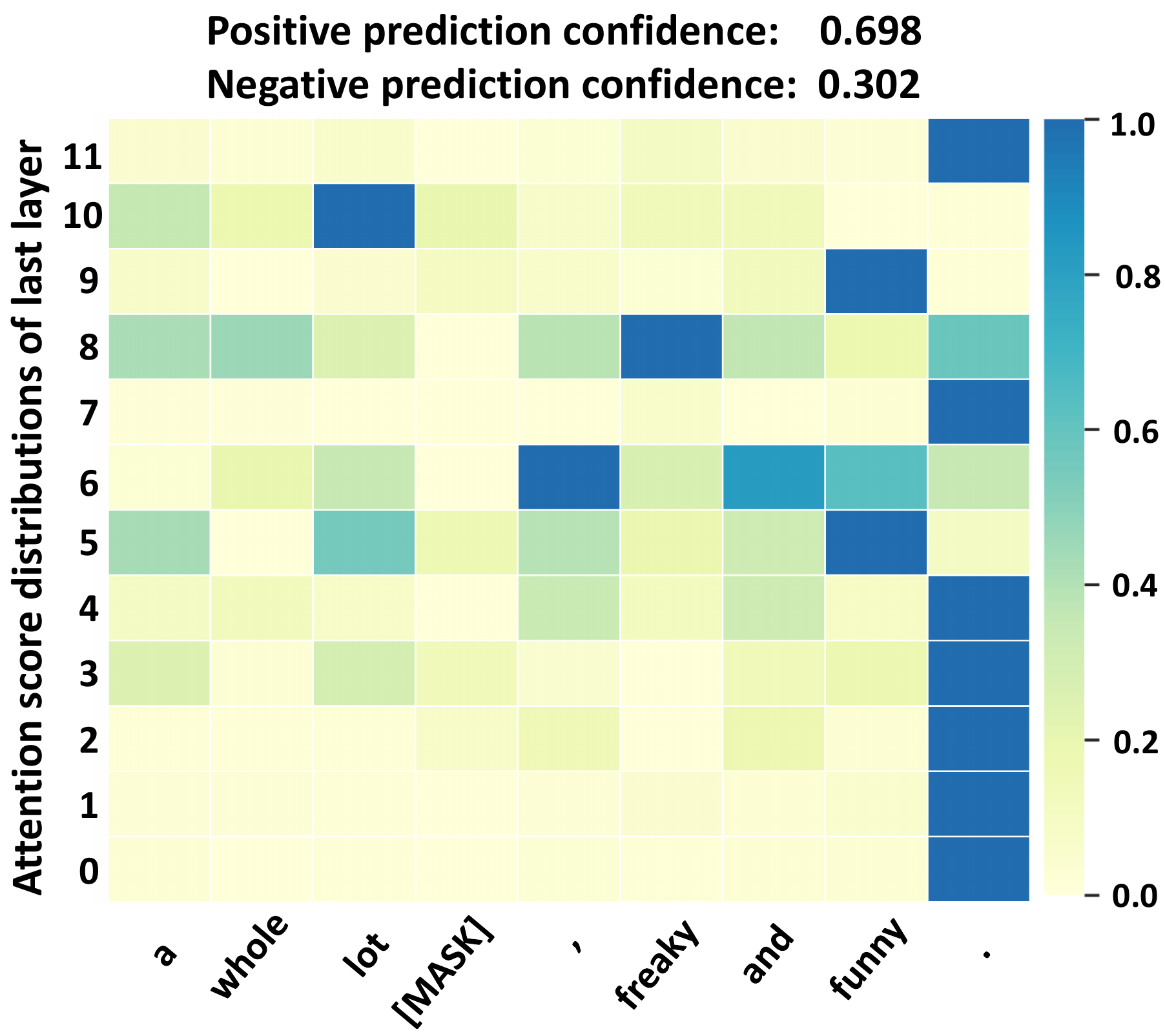}
        \label{fig:after_mask}
    \end{minipage}%
    }%
	\caption{
	The multi-head attention scores of each word on the last layer, obtained by BERT on SST dataset. 
	The ground-truth of sentence ``a whole lot foul, freaky and funny.'' is positive.}
	\label{fig:intro_case}
\end{figure*}

\paragraph*{Motivating Example: One Symptom of Overfitting} This paper is motivated by the observation that many fine-tuned models are very sensitive to irrelevant or misleading words in a sentence. 
For example, consider the sentence ``a whole lot foul, freaky and funny.'' from SST (Stanford Sentiment Treebank) dataset. 
In vanilla BERT, this sentence is predicted as negative while its actual label is positive. 
As illustrated in Figure \ref{fig:original}, one reason behind this may be that the word `foul', which is often associated with negative predictions, was given too predominant attention score by the BERT model. Instead, by masking the misleading word `foul' (see Figure \ref{fig:after_mask}), BERT is able to focus on the more relevant word `funny', and the final result flips to the correct label. 

\paragraph*{Self-Supervised Attention (SSA)}
In this paper, we hope to alleviate this above problem by introducing auxiliary knowledge to the attention layer of BERT. 
An interesting aspect of our approach is that we do not need any additional data or annotation (such as those in MT-DNN~\cite{liu2019multi}) for this auxiliary task. 
Instead, we propose a novel mechanism called self-supervised attention (SSA), which utilizes self-supervised information as the auxiliary loss. 

The idea behind SSA is simple --- Given a sentence of $n$ tokens $S = (t_1,...t_i...,t_n)$ predicted by the model with label $y$, we change the sentence into $S'$ by, for example, masking a token $t_i$, and generate the predicted label as $y'$. 
If $y'$ is different from $y$, we set the SSA score of $t_i$ as 1, otherwise 0.
Intuitively, this means that if the token $t_i$ plays a prominent role in the current model (i.e., the label prediction will be flipped if this token is masked), it will correspond to higher importance with a larger SSA score. Therefore, we can leverage a SSA layer to improve the accuracy and generalization ability of the original model by correctly predicting each token’s SSA score.

At first glance, it might be mysterious how the SSA score helps the performance of a deep neural network model --- all the information of SSA comes from the model itself. The intuition behind our approach is to impose constraints over the model to obtain several good properties. (1) If a model can predict SSA scores correctly, its decision surface is relatively smooth and less noisy. In other words, if we randomly mask some irrelevant or misleading words, the decision surface will become more robust. (2) By forcing the learned features to predict the auxiliary task of SSA, the learned features are able to encode more information about the keywords in a sentence and obtain the better capability of generalization.

\paragraph*{Summary of Technical Contributions}
Our first contribution is a co-training framework that precisely encodes the above motivation by a joint loss of downstream task and SSA. 
This itself already improves the performance of BERT significantly on some tasks. 
We then explore the idea of directly integrating SSA as an attention layer into the model. We propose a hybrid framework that combines both co-training and an additional self-supervised attention layer as our second contribution.
This further improves the performance of our model on a variety of public datasets.
We also demonstrate that the token-level SSA scores learned by the model agree well with human common-sense. 

\begin{figure*}[ht]
	\centering
	\includegraphics[width=0.8\linewidth]{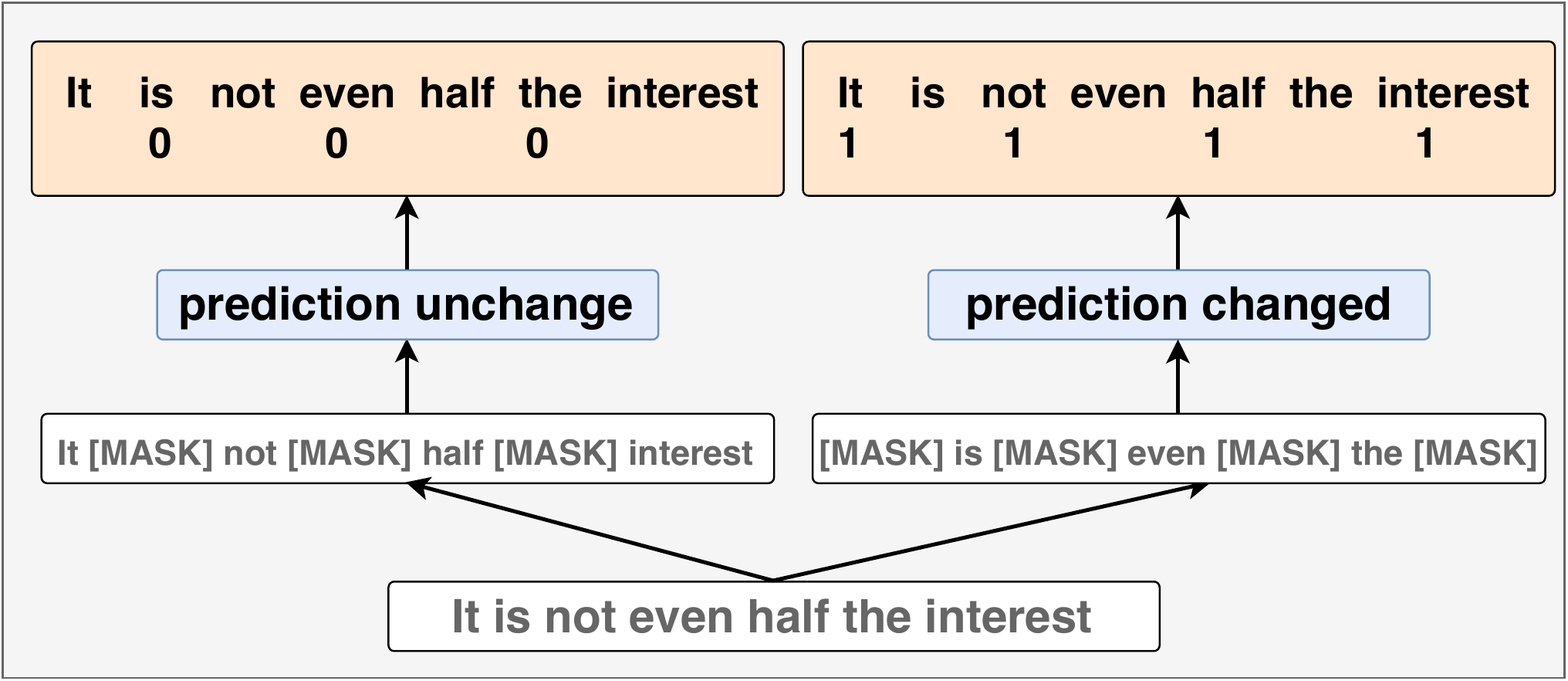}
	\caption{Label generation procedure}
	\label{fig:generate_label}
\end{figure*} 

\section{Related Work}
One way of addressing the overfitting problem is to increase the volume of training data. 
STILTs~\cite{phang2018sentence} introduces an intermediate-training phase rather than directly fine-tuning on the target task, for example, training on MNLI dataset before fine-tuning on the target RTE dataset. 
RoBERTa~\cite{liu2019roberta} focuses on training a better language model, which obtains a more powerful pre-trained checkpoint and improves the performance of downstream tasks significantly. Specifically, it collects a larger corpus, employs better hyper-parameters, and trains more iterations than BERT.
\cite{wu2019conditional} proposes a data augmentation method, named conditional BERT contextual augmentation, to generate synthetic data samples to be additional training data. Similarly, \cite{wei2019eda} explores four simple text editing techniques for enlarging the volume of data and improves the performance of CNN/RNN models. However, it is also noted that the improvement of such data augmentation methods is negligible when using pre-trained models like ELMo and BERT.

Another line of research focuses on auxiliary tasks. For instance, \cite{sun2019utilizing} leverages powerful BERT to tackle Aspect-Based Sentiment Analysis (ABSA) task. By constructing an auxiliary sentence from the aspect, authors convert ABSA to a sentence-pair classification task, such as Question Answering (QA) and Natural Language Inference (NLI). In this way, the original ABSA task is adapted to BERT, and new state-of-the-art results are established. MT-DNN~\cite{liu2019multi} presents a multi-task learning framework that leverages large amount of cross-task data samples to generate more general representations. AUTOSEM~\cite{guo2019autosem} proposes a two-stage multi-task learning pipeline, where the most useful auxiliary tasks are selected in the first stage and the mixing ratios of the tasks are learned in the second stage.
\cite{swayamdipta2018syntactic} incorporates syntactic information as an auxiliary objective and achieves competitive performance on the task of semantic role labeling.

This work also relates to the attention mechanism, which has been extensively utilized for modeling token relationships in natural languages. Previous works mainly focus on how to enhance sentence representation. For instance, \cite{yang2016hierarchical} proposes hierarchical attention network to progressively build a document vector for classification; \cite{lin2017structured} extracts an interpretable sentence embedding in a 2-D matrix through self-attention;
\cite{chen2018enhancing} utilizes a vector-based multi-head attention pooling method to enhance sentence embedding. \cite{tan2017abstractive} is the most relevant method to our work, which utilizes a graph-based attention mechanism in a sequence-to-sequence framework for document summarization. 

In addition, some methods of interpretability are relevant to the SSA score calculation strategy. For example, LIME~\cite{ribeiro2016should} focuses on explaining model's predictions by sampling perturbations of inputs and treating decision flipping as a key to explanation.
Besides, the gradient-based techniques such as GradCAM~\cite{selvaraju2017grad} are also applicable for SSA score calculation. 
The major focus of this paper is to address the usefulness of token importance to improve model accuracy rather than model interpretability.
We note that the SSA score generation strategy may be improved by LIME or GradCAM in the future work. 

The method proposed in this paper is different from previous works in the following perspectives: (1) Our solution does not require any extra knowledge (such as WordNet~\cite{wei2019eda} and back-translation~\cite{xie2020unsupervised}) to build augmented dataset, Instead, we explore the token-level importance as an auxiliary task and utilize it to improve the generalization ability of the model. (2) We propose a novel auxiliary task, Self-Supervised Attention (SSA), which differentiates the importance of each token with respect to the optimization target. (3) We apply an additional self-supervised attention layer to BERT for amplifying the effect of relevant tokens and diminishing the impact of irrelevant or misleading tokens. With the self-supervised attention layer, the BERT model becomes more resistant to overfitting in the fine-tuning stage.

\section{Model} \label{method}

This section presents an overview of the proposed \textit{Self-Supervised Attention} (SSA) model.
In Section \ref{SSA}, we introduce the auxiliary task of SSA and describe the training data generation procedure for this task. 
Then, in Section \ref{cotraining}, we introduce the co-training framework, which simultaneously optimizes the target task and auxiliary task. Finally, in Section \ref{hybridl}, we propose the hybrid model with self-supervised attention layer. 

\subsection{The auxiliary task of SSA}
\label{SSA}
Based on a sentence $S$, we generate another sentence $S'$ by masking several tokens chosen at random in the original sentence. Sentence $S'$ can be seen as a noisy counterpart of $S$.
At a high level, by examining the relationship between $S$ and $S'$, we can construct a smoother surface passing through both $S$ and $S'$, and thus allow more robust local minimum point to be reached via optimization. 
Specifically, if $S'$ has the same prediction label as $S$, the modified tokens can be de-emphasized for improving the generalization capability of the original model. Otherwise, we want to emphasize the modified tokens to address their task-oriented importance.
Motivated by this, we propose a novel auxiliary task: \textit{Self-Supervised Attention} (SSA), which learns a task-oriented weight for each token. When we correctly predict the weights of the tokens for a specific task, the decision surface between $S$ and $S'$ will be smoother and so, the model will have a better generalization capability. 

Given a sentence $S = (t_1,...t_i...,t_n)$, the SSA task outputs a binary output vector $Y = (y_1^t,...,y_i^t...,y_n^t)$, where $y_i^t$ indicates how important the token $t_i$ is to the target task. 
The loss function can be formulated as follows: 

\begin{equation}
    \mathcal{L}_{SSA}= -\sum_{i}\ell_{SSA}(y_{i}, y_{i}^t)\label{eq:l_ssa}
\end{equation}

\begin{equation}
    y_i^t=\sigma \big(w_{SSA}^i M(t_{i}) \big)\label{eq:y_ssa}
\end{equation}

\noindent where $M(t_{i})$ denotes the model from the previous epoch, $w_{SSA}^i$ is the fully connected layer of the $i$-th token for SSA task, $\sigma$ is a softmax operation, $y_i$ denotes the SSA label of token $t_i$, and $\ell_{SSA}$ is the cross entropy loss.

The training data generation procedure for SSA task is described as follows. Given a sentence $S$, we first mask several tokens randomly and get a new sentence $S'$. Then, we inference the predicted label of two sentences using the same model, $M(t_{i})$. If both predicted labels are the same, we set the importance labels of masked tokens as 0, as they have little impact on the target task. Otherwise, these tokens should be emphasized and corresponding importance labels are set to be 1.
We only process the sentences $S$ that the prediction of $M(t_{i})$ is correct to get rid of noises.
The benefit of this operation is that, with the increase of epochs, the label given by $M(t_{i})$ becomes more and more accurate.
Besides, the number of tokens being masked is proportional to the length of the sentence, which is set to $0.3$ empirically.
We use a generation ratio $\gamma$ to control the amount of generated sentences, and larger $\gamma$ means more sentences will be generated.

We train different samples generated by the same sentence in different iterations without slowing down the training procedure. Figure \ref{fig:generate_label} shows two training samples generated by a sentence in multiple iterations.
Suppose that we have an input sentence $S=$``It is not even half the interest'', we generate the masked sentence $S'_1=$``It [MASK] not [MASK] half [MASK] interest'' or $S'_2=$``[MASK] is [MASK] even [MASK] the [MASK]''. $S'_1$ does not change the prediction label, so the masked tokens are labeled by 0, that is, $Y_1=(\_, 0, \_, 0, \_, 0, \_)$, where ``$\_$'' does not contribute to the loss function. On the other hand, $S'_2$ flips the original prediction of $S$, so we have $Y_2=(1, \_, 1, \_, 1, \_, 1)$. 

\subsection{Co-training Framework}
\label{cotraining}

\begin{figure}[ht]
	\centering
    \includegraphics[width=4.5cm]{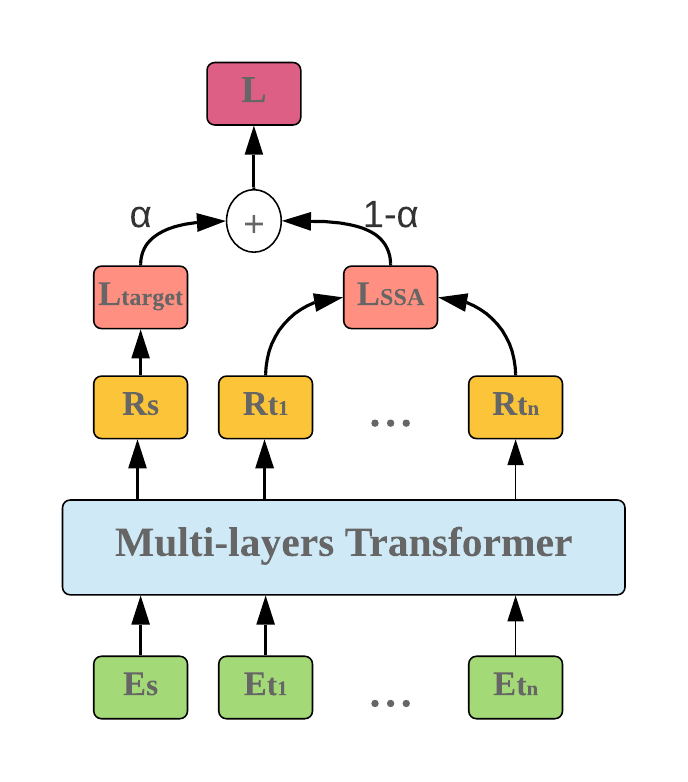}
	\caption{The model architecture for co-training}
	\label{fig:co-training}
\end{figure}
A straight-forward way to leverage the SSA task is training the target task and SSA jointly. 
Specifically, the overall loss can be defined as a linear combination of two parts: 

\begin{equation}
    \mathcal{L} =\alpha \mathcal{L}_{target} + (1-\alpha) \mathcal{L}_{SSA}\label{eq:L}
\end{equation}

\begin{equation}
     \mathcal{L}_{target} = \sum_{i=1}\ell_{target}(y_{i}, y_{i}^{s})\label{eq:L_target}
\end{equation}

\noindent where $\ell_{target}$ is the loss function of the target task, for example, negative log-likelihood loss for sentiment classification and mean-squared loss for regression task; $y_{i}$ and $y_i^{s}$ denote the actual label of sentence $s_i$ and the predicted label for the target task respectively; $\mathcal{L}_{target}$ denotes the loss function of the target task while $\mathcal{L}_{SSA}$ represents for that of SSA task; $\alpha$ is a linear combination ratio which controls the relative importance of two losses.

The model architecture for co-training is illustrated in Figure \ref{fig:co-training}. Each token in the input sentence is mapped to a representation, and then fed into an encoder of BERT or multi-layers Transformer. The output of the encoder consists of a sentence representation vector $R_s$ and token-level representation vectors $[R_{t_1}, R_{t_2}, ..., R_{t_n}]$. The sentence representation is used for target task prediction and the token representation vectors are leveraged in the SSA task.  The co-training framework optimizes these two tasks by the target loss and auxiliary loss of SSA alternatively. The pseudo code is described in Algorithm~\ref{alg:Joint-training}.

\begin{algorithm}[ht]
\caption{Co-training Procedure} 
\SetKwInOut{Input}{input}
\SetKwInOut{Output}{output}
\Input{ target data $D$, training epochs $T$}
\Output{ joint model $M_t$}
fine-tune BERT on target task, get $BERT_0$\\
$t \gets 0$\\
init $M_t$ $\gets$ $BERT_0$\\
\While{ $t < T$ } {
    leverage $M_t$ to generate SSA labels, get $D_t$ \\
    fine-tune $M_{t}$ on target data $D_t$\\
    init $M_{t+1}$ $\gets$ $M_t$\\
    $t \gets t + 1$
}
\Return{ $M_t$ }
\label{alg:Joint-training}
\end{algorithm}

\subsection{Hybrid Model with SSA layer}
\label{hybridl}

\begin{figure}[ht]
	\centering
    \includegraphics[width=4.5cm]{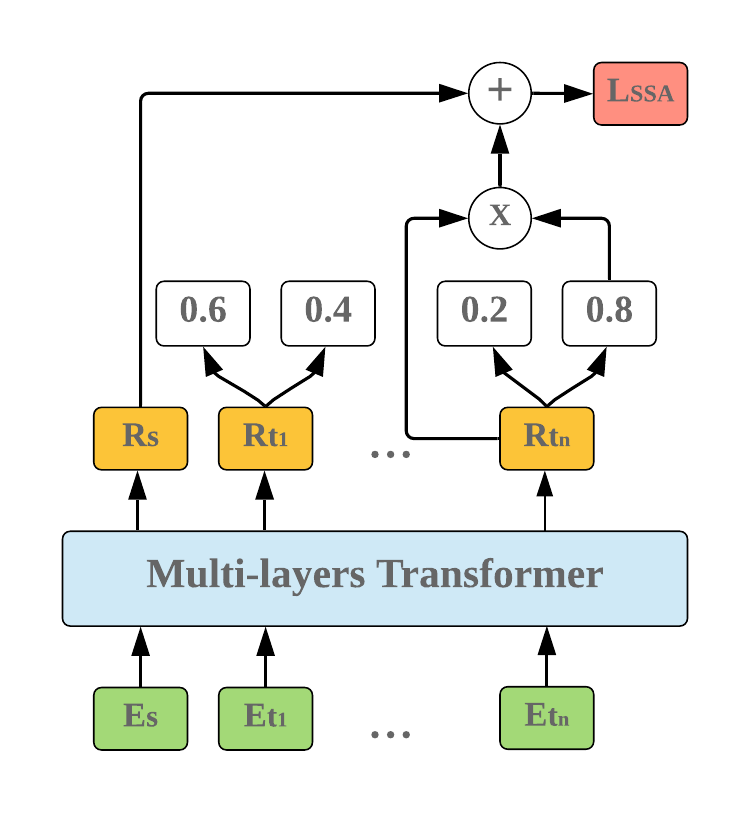}
	\caption{Hybrid Model with SSA layer}
	\label{fig:attention}
\end{figure}

\begin{figure*}[!t]
	\centering
	\includegraphics[width=0.85\linewidth]{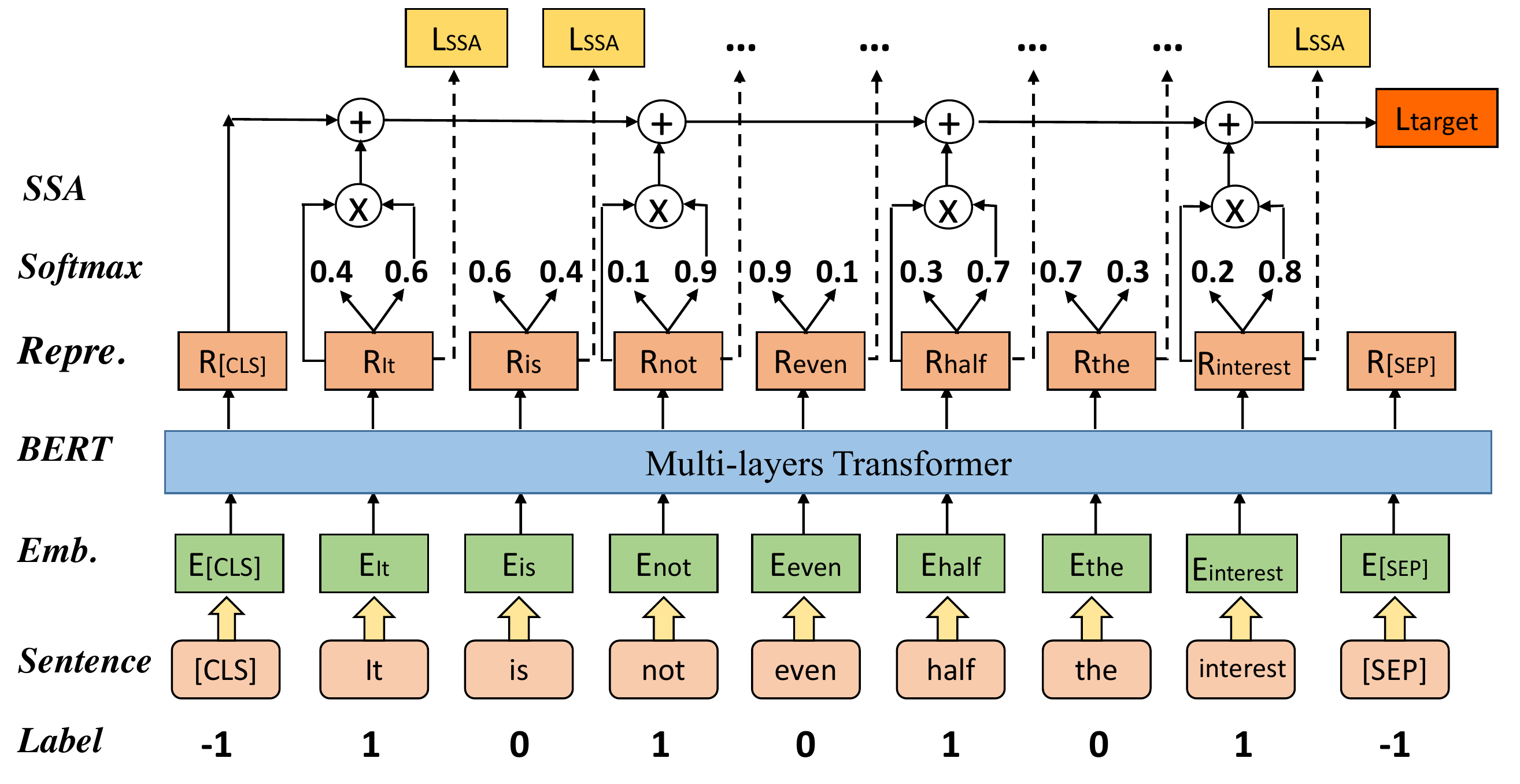}
	\caption{Illustration of an example based on the Hybrid model}
	\label{fig:model}
\end{figure*}

The limitation of co-training framework is that the SSA task can not impact target prediction explicitly. It only acts as a regularizer on the loss function to force the learned embedding vectors of tokens to encode their relevant importance. Intuitively, if an irrelevant or misleading token exists in an training sentence, we can mask the token explicitly and guide the model to capture more important information for alleviating the overfitting problem. Therefore, we add an additional \textit{self-supervised attention} (SSA) layer on the vanilla BERT model, which readjusts the weight of each token explicitly according to their relative importance to the target task.

\begin{equation}
    R_o=\beta \cdot R_{[CLS]} + (1-\beta) \sum_{i}\big(\sigma(y_i^t)\cdot R_i \big)
    \label{hybrid}
\end{equation}

The hybrid model with SSA layer is illustrated in Figure \ref{fig:attention} and Equation \ref{hybrid} is the mathematical formulation. 
It yields an extra sentence representation by summing up all the token embedding vectors $R_i$ weighted by the SSA prediction scores $y_i^t$ after the softmax operation $\sigma$. 
Afterwards, the extra sentence embedding output by the SSA layer and the original sentence embedding vector $R_{[CLS]}$ are linearly combined as the final sentence representation $R_o$. $\beta$ is a hyper-parameter to control the relative weight of this linear combination. 

Figure \ref{fig:model} further demonstrates a concrete example of hybrid model. 
Take ``It is not even half the interest'' as input, the discrete tokens are mapped to embedding representations first, then the embedding vectors are transformed by a multi-layer transformer encoder (e.g., BERT), to generate sentence embedding and token representations.
As a result, the SSA layer identifies that `It', `not', `half', and `interest' are important for the target sentiment classification task, and then constructs an extra sentence embedding by weighted summation of token-level embedding vectors based on the corresponding SSA scores. The final sentence representation is a linear combination of original sentence embedding and extra sentence embedding. The loss function of primary task and SSA task are jointly optimized in a co-training framework.

As with the co-training framework, the model training step and SSA data generation are alternatively processed. The detailed training procedure is identical to algorithm \ref{alg:Joint-training} except that we train the hybrid model instead of the joint model.

\begin{table*}[ht]
    \renewcommand\arraystretch{1.0}
    \centering
    \begin{tabular}{cccccc}
    \hline
     Task & \#Train & \#Dev & \#Test & C. & Description  \\ \hline
     CoLA & 8,551 & 1,042 & 1,064 & 2 & Corpus of Linguistic Acceptability \\
     SST-2 & 67,350 & 873 & 1,822 & 2 & Stanford Sentiment Treebank   \\
     MRPC & 3,669 & 409 & 1,726 & 2 & Whether the sentence pair are semantically equivalent \\
     STS-B & 5,750 & 1,501 & 1,380 & * &  How similar two sentences are in semantic meaning\\
     QQP & 363,871 & 40,432 & 390,965 & 2 & Whether two questions are semantically equivalent \\
     MNLI & 392,703 & 9,816/9,833 & 9,797/9,848 & 3 & Multi-Genre Natural Language Inference \\
     QNLI & 104,744 & 5,464 & 5,464 & 2 & Question Natural Language Inference  \\
     RTE & 2,491 & 278 & 3,001 & 2 & Recognizing Textual Entailment (like MNLI)\\
     WNLI & 636 & 72 & 147 & 2 & Small natural language inference dataset \\
     \hline
     SentiHood & 45,025 & 11,245 & 22,549 & 2 & Targeted Aspect-based sentiment analysis dataset (TABSA) \\
     SemEval-2014 & 68,490 & 7,610 & 20,000 & 2 & Aspect-based sentiment analysis dataset (ABSA) \\
     \hline
    \end{tabular}
    \caption{Statistics of all datasets. The numbers on the left and right side of character ``/'' for task MNLI represent MNLI-m and MNLI-mm correspondingly; ``\#C.'' is number of categories in the task; ``*'' denotes regression task.}
    \label{tab:Statistics}
\end{table*}

\section{Experiments}
In this section, we describe the experimental details about the proposed models \textbf{SSA-Co} and \textbf{SSA-Hybrid} on the GLUE benchmark and ABSA tasks.

\subsection{Tasks}
The General Language Understanding Evaluation (GLUE) benchmark is a collection of datasets for training and evaluating neural language understanding models. 
The statistics of datasets are summarized in Table \ref{tab:Statistics}, where \textbf{CoLA}~\cite{warstadt2019neural} and \textbf{SST-2}~\cite{socher2013recursive} are datasets for single-sentence classification tasks, \textbf{STS-B}~\cite{cer2017semeval} is for text similarity task, \textbf{MRPC}~\cite{dolan2005automatically} and \textbf{QQP} are binary classification tasks, and the rests are natural language inference tasks including \textbf{MNLI}
~\cite{williams2018broad}, \textbf{QNLI}~\cite{rajpurkar2016squad}, \textbf{RTE}~\cite{dagan2005pascal} and \textbf{WNLI}~\cite{levesque2012winograd}. We follow the default evaluation metrics for tasks in GLUE benchmark. 

For each task, we use the default train/dev/test split. The model is trained on the training set, while the hyper-parameters are chosen based on the development set. We submit the prediction results of test set to the GLUE evaluation service\footnote{https://gluebenchmark.com} and report the final evaluation scores. 

To further evaluate the effectiveness of the proposed models, we conduct experiments on the aspect-based sentiment analysis (ABSA) task. ABSA task is more than a pure sentiment polarity classification task, it also involves aspect detection. We argue that this complex task can also benefit from the proposed SSA task. We follow~\cite{sun2019utilizing} to conduct experiments on two ABSA task datasets: SentiHood~\cite{saeidi2016sentihood} and SemEval-2014 Task 4~\cite{saeidi2016sentihood}. Note that for SemEval-2014 Task 4, we jointly evaluate subtask 3 (Aspect Category Detection) and subtask 4 (Aspect Category Polarity). Since SemEval-2014 task 4 only provides train/test split, we randomly select 10\% samples from the training set as validation data. For SentiHood we follow the default train/dev/test split. The detailed statistics of these two datasets are also listed in Table \ref{tab:Statistics}. 

\subsection{Algorithms}
The pre-training and fine-tuning frameworks have evolved rapidly since BERT was proposed.
As introduced in the Related Work section, there are many existing works on data augmentation and leveraging auxiliary tasks. Our proposed algorithm is orthogonal to these techniques and can be applied to more advanced models. In this paper, the comparison is mainly on the basis of vanilla BERT while more experiments on other model variants (e.g., MT-DNN) are left to future works. 

The configurations of baselines and our solutions are described as below. 

\textbf{BERT}~\cite{devlin2019bert} is a multi-layer bidirectional transformers. It is first pre-trained on a large corpus containing 3,300M words, and then fine-tuned on downstream tasks. We download the checkpoint of BERT-base\footnote{\url{https://storage.googleapis.com/bert_models/2018_10_18/uncased_L-12_H-768_A-12.zip}} and BERT-large\footnote{\url{https://storage.googleapis.com/bert_models/2018_10_18/uncased_L-24_H-1024_A-16.zip}} from the official website.

\textbf{BERT-mask} is a simple baseline to our algorithm, where the training data is augmented by randomly masking tokens in the original sentences in the same way as SSA. By comparing our solution with this simple baseline, we can clearly examine the superiority of self-supervised attention.

\textbf{BERT-EDA}~\cite{wei2019eda} is another baseline to our algorithm, where training sentences are edited by four operations: synonym replacement, random insertion, random swap and random deletion. We increase the data volume as large as other models use. This model utilizes the WordNet as the guidance, while ours only extracts the token importance information from the original data.

\textbf{BERT-ABSA}~\cite{sun2019utilizing} enhances BERT by constructing an auxiliary sentence from the aspect and converts the ABSA task to the sentence-pair classification task, which is designed for SentiHood and SemEval2014 Task 4 datasets. We compare with BERT-single (original data format) and BERT-pair (sentence-pair data format) models proposed in this paper. Correspondingly, for the ABSA task we name our SSA-based model as BERT-single-H and BERT-pair-H.

\textbf{RoBERTa}~\cite{liu2019roberta} is an improved recipe of BERT. 
It is worth noting that the results reported on the leaderboard are ensembles of single-task models.
For a fair comparison, the results of RoBERTa in our experiments are reproduced based on the officially released checkpoint\footnote{\url{https://github.com/pytorch/fairseq}}. 

\textbf{SSA-Co} and \textbf{SSA-Hybrid} are the models proposed in this paper, where SSA-Co takes SSA as an auxiliary task and leverages the co-training framework to optimize the two tasks jointly; SSA-Hybrid (or SSA-H) takes the hybrid model which contains an additional \textit{self-supervised attention} (SSA) layer in the network structure.

\begin{table*}[!t]
    \small
    \renewcommand\arraystretch{1.2}
    \setlength{\abovecaptionskip}{5pt}
    \begin{tabular}{lllllllllll}
    \hline
    \multicolumn{11}{c}{\textbf{\normalsize{GLUE test set results of BERT-base}}} \\ \hline
     Model & Avg & CoLA & SST-2 & MRPC & STS-B & QQP & MNLI & QNLI & RTE & WNLI  \\ \hline
     BERT & 78.3 & 52.1 & 93.5 & 88.9/84.8 & 87.1/85.8 & 71.2/89.2 & 84.6/83.4 & 90.5 & 66.4 & \textbf{65.1} \\ 
     BERT-r & 78.4 & 51.8 & 94.0 & 87.1/82.8 & 86.9/86.0 & 71.0/89.1 & 84.2/83.2 & 90.5 & 69.1 & \textbf{65.1} \\ 
     BERT-m & 78.3 & 53.1 & 93.7 & 88.1/83.9 & 86.6/85.4 & 69.8/88.6 & 84.2/83.1 & 90.6 & 67.2 & \textbf{65.1} \\
     BERT-EDA & 78.4 & 52.1 & 93.6 & 88.6/84.4 & 86.8/85.6 & 70.5/88.3 & 84.2/83.2 & 90.5 & 68.9 & \textbf{65.1} \\
     \hline
     BERT-SSA-Co & 79.0 & 52.8$^*$ & 94.1 & 89.0/85.2$^*$ & 87.2/86.1 & 71.1/89.1 & 84.6/83.7$^*$ & 90.8 & 69.8$^{**}$ & \textbf{65.1} \\ 
     BERT-SSA-H & \textbf{79.3} & \textbf{54.1}$^*$ & \textbf{94.4} & \textbf{89.1/85.4}$^*$ & \textbf{87.2/86.3} & \textbf{71.5/89.3}$^{**}$ & \textbf{84.7/84.1}$^*$ & \textbf{91.3}$^*$ & \textbf{70.3}$^{**}$ & \textbf{65.1} \\ 
     \hline
    \end{tabular}
    \caption{GLUE test set results scored by GLUE evaluation service. The results of BERT-r are reproduced with the open sourced codes and the reproduction numbers are on-par with the numbers reported in the GLUE leaderboard.
    ``BERT-m'' represents our baseline model ``BERT-mask''; ``BERT-EDA'' is another baseline model proposed by~\cite{wei2019eda};
    ``BERT-SSA-Co'' and ``BERT-SSA-H'' denote our SSA-Co and SSA-Hybrid method respectively; The result on the left and right side of character ``/'' for task MNLI represents MNLI-m and MNLI-mm correspondingly;
    We also conduct significant test between BERT-SSA models and BERT, where * means the improvement over vanilla BERT is significant at the 0.05 significance level; ** means the improvement over vanilla BERT is significant at the 0.01 significance level.}
    \label{tab:glue_test}
\end{table*}

\begin{table*}[ht]
    \small
    \renewcommand\arraystretch{1.2}
    \setlength{\abovecaptionskip}{5pt}
    \begin{tabular}{lllllllllll}
    \hline
     \multicolumn{11}{c}{\textbf{\normalsize{GLUE test set results of BERT-large}}} \\ \hline
     Model & Avg & CoLA & SST-2 & MRPC & STS-B & QQP & MNLI & QNLI & RTE & WNLI  \\ \hline
     BERT & 80.5 & 60.5 & 94.9 & 89.3/85.4 & 87.6/86.5 & 72.1/89.3 & 86.7/85.9 & 92.7 & 70.1 & \textbf{65.1}  \\
     BERT-r & 80.5 & 60.2 & 94.9 & 89.2/85.3 & 87.6/86.5 & 72.1/89.4 & 86.5/85.9 & 92.8 & 70.1 & \textbf{65.1}   \\ \hline
     BERT-SSA-H & \textbf{81.2} & \textbf{62.7}$^{**}$ & \textbf{95.8}$^{*}$ & \textbf{90.2/86.2}$^{*}$ & \textbf{88.1/87.0}$^{*}$ & \textbf{72.4/89.5} & \textbf{87.0}/\textbf{86.3}$^{*}$ & \textbf{92.9} & \textbf{71.5}$^{**}$ & \textbf{65.1} \\ \hline \hline
     
     \multicolumn{11}{c}{\textbf{\normalsize{GLUE test set results of RoBERTa-base}}} \\ \hline
     Model & Avg & CoLA & SST-2 & MRPC & STS-B & QQP & MNLI & QNLI & RTE & WNLI  \\ \hline
     RoBERTa-r & 81.5 & 61.1 & 95.6 & 91.3/88.5 & 88.1/87.2 & 72.2/89.4 & 87.1/86.3 & 92.8 & 73.9 & \textbf{65.1} \\ \hline
     RoBERTa-SSA-H & \textbf{82.2} & \textbf{62.0}$^{*}$ & \textbf{96.2}$^{*}$ & \textbf{92.2/89.3} & \textbf{89.6/88.7}$^{*}$ & \textbf{72.5/89.6} & \textbf{87.4/87.0}$^{*}$ & \textbf{93.1} & \textbf{74.9}$^{*}$ & \textbf{65.1} \\ 
     \hline  \hline
     
     \multicolumn{11}{c}{\textbf{\normalsize{GLUE test set results of RoBERTa-large}}} \\ \hline
     Model & Avg & CoLA & SST-2 & MRPC & STS-B & QQP & MNLI & QNLI & RTE & WNLI  \\ \hline
     RoBERTa-r & 83.8 & 63.9 & 96.3 & 91.5/88.3 & 91.6/91.1 & 72.6/89.7 & 89.5/\textbf{89.6} & 94.3 & 82.3 & \textbf{65.1} \\ \hline
     RoBERTa-SSA-H & \textbf{84.3} & \textbf{64.9}$^{*}$ & \textbf{96.8}$^{*}$ & \textbf{92.6/90.2}$^{*}$ & \textbf{91.8/91.2} & \textbf{73.5/90.0}$^{*}$ & \textbf{89.8}/\textbf{89.6} & \textbf{94.6} & \textbf{82.9}$^{*}$ & \textbf{65.1} \\ 
     \hline 
    \end{tabular}
    \caption{The GLUE results of our model based on BERT-large and RoBERTa. We also conduct significant test between SSA-H enhanced model and vanilla pre-trained model, where * means the improvement over vanilla model is significant at the 0.05 significance level; ** means the improvement over vanilla model is significant at the 0.01 significance level.}
    \label{tab:glue_test_2}
\end{table*}

\subsection{Evaluation Metrics}
During the evaluation of GLUE benchmark, following~\cite{wang2018glue}, we use different evaluation metrics for different dataset. For CoLA, we use MCC (Matthews correlation coefficient) score; Regression task STS-B is person correlation coefficient and spearman correlation coefficient; MRPC and QQP are accuracy and F1 score; And others are only accuracy score.
For SentiHood task, we consider four aspects which appear the most frequently (general, price, transit location and safety)~\cite{saeidi2016sentihood}. The evaluation metrics for aspect detection are strict accuracy, Macro-F1 and AUC, while those for sentiment classification are accuracy and macro-average AUC.
For Semeval-2014 task 4, we use Precision, Recall, and Micro-F1 to evaluate subtask 3 (aspect category detection), and use accuracy score for subtask 4 (aspect category polarity)~\cite{pontiki2016semeval}. 4-way, 3-way and 2-way settings refer to how many categories should be included in the calculation process.

\subsection{Detailed Settings}
All experiments are based on the publicly available implementation of BERT and RoBERTa (PyTorch)\footnote{https://github.com/huggingface/pytorch-pretrained-BERT}. For GLUE tasks, we follow the fine-tuning regime specified in~\cite{devlin2019bert}. 
While many submissions to the GLUE leaderboard\footnote{https://gluebenchmark.com/leaderboard} depend on multitask fine-tuning or ensembles, \textbf{our submission depends only on single-task fine-tuning}.

We employ grid search to find the optimal hyper-parameters according to a pre-defined range: learning rate $lr\in$ \{1e-5, 2e-5\}, batch size $b\in$ \{16, 32\}, epochs $T\in$ \{2, 3, 5\}, loss combination ratio $\alpha\in$\{0.7, 0.9\}, linear combination ratio $\beta\in$\{0.2, 0.5, 0.9\} and auxiliary data generation ratio for SSA and BERT-m $\gamma \in \{0.6, 1.0, 2.0\}$. 
The maximum sequence length is set to 512 in all of our experiments and we will cut the extra length in each batch to accelerate the training speed.

Further, we evenly split the validation set into three subsets when fine-tuning BERT models and the hyper-parameters achieving high average performance with the smallest variance on the three validation subset will be chosen.

\subsection{GLUE Results}

The experimental results on GLUE benchmark are summarized in Table \ref{tab:glue_test} and Table \ref{tab:glue_test_2}. It is a known problem that the train/dev split for WNLI is correct but somewhat adversarial, and many papers exclude WNLI in their tables, including the original paper of BERT~\cite{devlin2019bert}. Therefore, we do not compare the performance on this dataset (only on the list for completeness). 

As shown in the Table \ref{tab:glue_test}, our methods consistently achieve better results than base BERT. Specifically, \textbf{SSA-Co} obtains better results on seven datasets, while keeps on-par performance on the rest of the datasets. It demonstrates that the auxiliary task of SSA is helpful for model generalization. 
Moreover, the hybrid model \textbf{SSA-Hybrid} performs better on all the datasets and pushes the average score of base BERT (reproduction version) from $78.4$ to $79.3$. 
The BERT-mask baseline also shows some advantages in several datasets, which can be explained by the effect of data augmentation.
However, the decay in other datasets (MRPC, STS-B, QQP and MNLI) indicates that such an improvement is unstable. The BERT-EDA baseline obtains similar results as BERT-mask, which verifies models that have been pre-trained on massive datasets with negligible improvement from easy data augmentation techniques.
Our model is obviously more robust, demonstrating the superiority of SSA-based hybrid model over models that only use data augmentation.

We also apply our solutions to more advanced baselines, i.e., BERT-large, RoBERTa-base, and RoBERTa-large. As shown in Table \ref{tab:glue_test_2}, \textbf{SSA-Hybrid} consistently outperforms other baselines on different datasets. For BERT-large, it wins on all datasets and leads to a $0.7$ absolute increase on the average score. The results further verify the advantages of self-supervised attention layer.
Moreover, \textbf{SSA-Hybrid} also achieves improvement on RoBERTa, especially on RoBERTa-base.
For RoBERTa-large, one of the strongest single models in the literature, this still leads to an absolute increase of $0.5$ on the average score.
Considering that no external knowledge is leveraged and only a few extra weights are introduced to the model, these improvements are significant.

In addition, we discuss gains on different data sizes. Among GLUE classification datasets, RTE, MRPC and CoLA have the smallest training data sizes. Our model respectively achieves 1.7\%, 2.7\% and 4.4\% relative lifts in these tasks compared to BERT-base, which is more significant than larger datasets like MNLI. This is reasonable because smaller datasets benefit more from generalization techniques.

\subsection{ABSA Results}

\begin{table}[htbp]
    \small
    \renewcommand\arraystretch{1.0}
    \centering
    \begin{tabular}{lccccc}
    \hline
    \multicolumn{6}{c}{\textbf{\normalsize{SentiHood test set results of BERT-base}}} \\ \hline
    \multirow{2}{*}{Model} & \multicolumn{3}{c}{\textbf{Aspect}} & \multicolumn{2}{c}{\textbf{Sentiment}}  \\  \cline{2-6}
      & Acc. & F1 & AUC & Acc. & AUC  \\ \hline
     BERT-single & 73.7 & 81.0 & 96.4 & 85.5 & 84.2 \\
     BERT-pair & 79.8 & 87.9 & 97.5 & 93.6 & 97.0 \\ \hline
     BERT-single-H & 75.3 & 82.7 & 96.6 & 86.3 & 85.3 \\
     BERT-pair-H & \textbf{82.0} & \textbf{88.2} & \textbf{97.9} & \textbf{94.0} & \textbf{97.4} \\ \hline
    \end{tabular}
    \caption{Base BERT performance on SentiHood dataset with the best performances bold. Top: ``BERT-single'' represents the baseline models fine-tuned on BERT with original datasets; ``BERT-pair'' is another baseline model proposed by~\cite{sun2019utilizing} trained with auxiliary Sentences. Bottom: ``BERT-single-H'' and ``BERT-pair-H'' denote the models incorporating our SSA layers.}
    \label{tab:ABSA_sentihood_base}
\end{table}

\begin{table}[htbp]
    \small
    \renewcommand\arraystretch{1.0}
    \centering
    \begin{tabular}{ccccccc}
    \hline
    \multicolumn{7}{c}{\textbf{\normalsize{Semeval-2014 test set results of BERT-base}}} \\ \hline
     Model & P & R & F1 & 4-w & 3-w & 2-w  \\ \hline
     BERT-single & 92.8 & 89.1 & 90.9 & 83.7 & 86.9 & 93.3 \\
     BERT-pair & 93.6 & 90.8 & 92.2 & 85.9 & 89.9 & 95.6 \\
      \hline
     BERT-single-H & 93.1 & 89.5 & 91.3 & 84.0$^{*}$ & 87.3$^{*}$ & 93.5$^{*}$ \\
     BERT-pair-H & \textbf{93.9} & \textbf{91.4} & \textbf{92.6} & \textbf{86.2}$^{*}$ & \textbf{90.3}$^{*}$ & \textbf{95.7}$^{*}$ \\ \hline
    
    \end{tabular}
    \caption{Base BERT performance on Semeval-2014 task 4 Subtask 3 and Subtask 4. ``n-w'' means the accuracy of n-way setting in Subtask 4. We also conduct significant test between SSA-Hybrid enhanced model and vanilla BERT in Subtask 4, where * means the improvement over vanilla model is significant at the 0.05 significance level.}
    \label{tab:ABSA_semeval_base}
\end{table}

Table \ref{tab:ABSA_sentihood_base} and \ref{tab:ABSA_semeval_base} present results of applying our approach to ABSA tasks and show that SSA-based BERT obtains the increased performance consistently both on the BERT-single and data-augmented BERT-pair configuration. Specifically, from the experiments on SentiHood, we learn that the proposed SSA can provide complementary information for pre-trained BERT. This means that, not only the sentiment classification task can benefit from our model, but also the aspect detection task. Across all the evaluation metrics, the \textbf{BERT-pair-H} combined with the SSA layer and data augmentation method performs the best.
\textbf{BERT-pair-H} obtains an 82.0\% accuracy for the aspect detection task, which is a 2.2\% absolute improvement compared to the original BERT-pair model. We also observe that the performance on the sentiment classification task is improved when using SSA hybrid model.
For Semeval-2014 task 4, the F1 scores are pushed forward substantially by at least 0.4\% compared to baseline models. It is noted that our proposed model achieves further performance based on the data augmentation of constructing auxiliary sentences~\cite{sun2019utilizing}. We conjecture that the ability of generalization learned by SSA plays a different role with existing data-augmented methods, which will be further studied.

\subsection{Training Cost Comparison}

\begin{figure}[ht]
	\centering
    \includegraphics[width=1.0\linewidth]{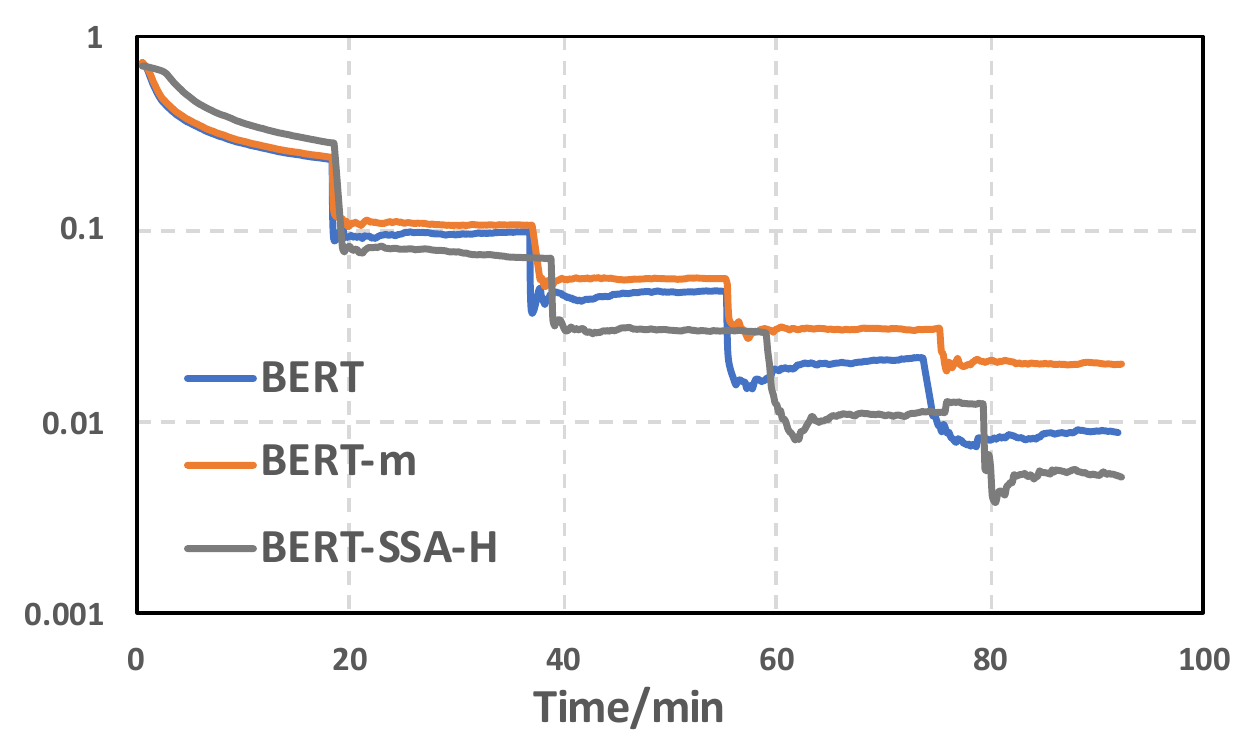}
	\caption{
	The x-axis represents training time and the y-axis represents the individual loss value of each epoch scaled using $\log_{10}$.}
	\label{fig:time}
\end{figure}

To further evaluate the extra training cost of SSA paid for the performance improvement, we analyze the training cost on SST-2 dataset. The results are demonstrated in Figure \ref{fig:time}. It is predictable that \textbf{BERT} and \textbf{BERT-m} converge faster than \textbf{BERT-SSA-H} in the initial iterations, since the SSA-based solution contains more parameters which lead to harder optimization. However, after one epoch, the SSA-based model begins to show its advantage and achieve a lower loss than baseline models with the same training time. 
Note that when the loss drops each time, the SSA-based model always has a short delay after two baselines, indicating the time cost of label generation. Nevertheless, as shown in the figure, this cost is minor compared with the total training time, and the final performance improvement justifies the cost. 
\subsection{Sensitivity Study}
\begin{figure}[ht]
	\centering
    \includegraphics[width=1.0\linewidth]{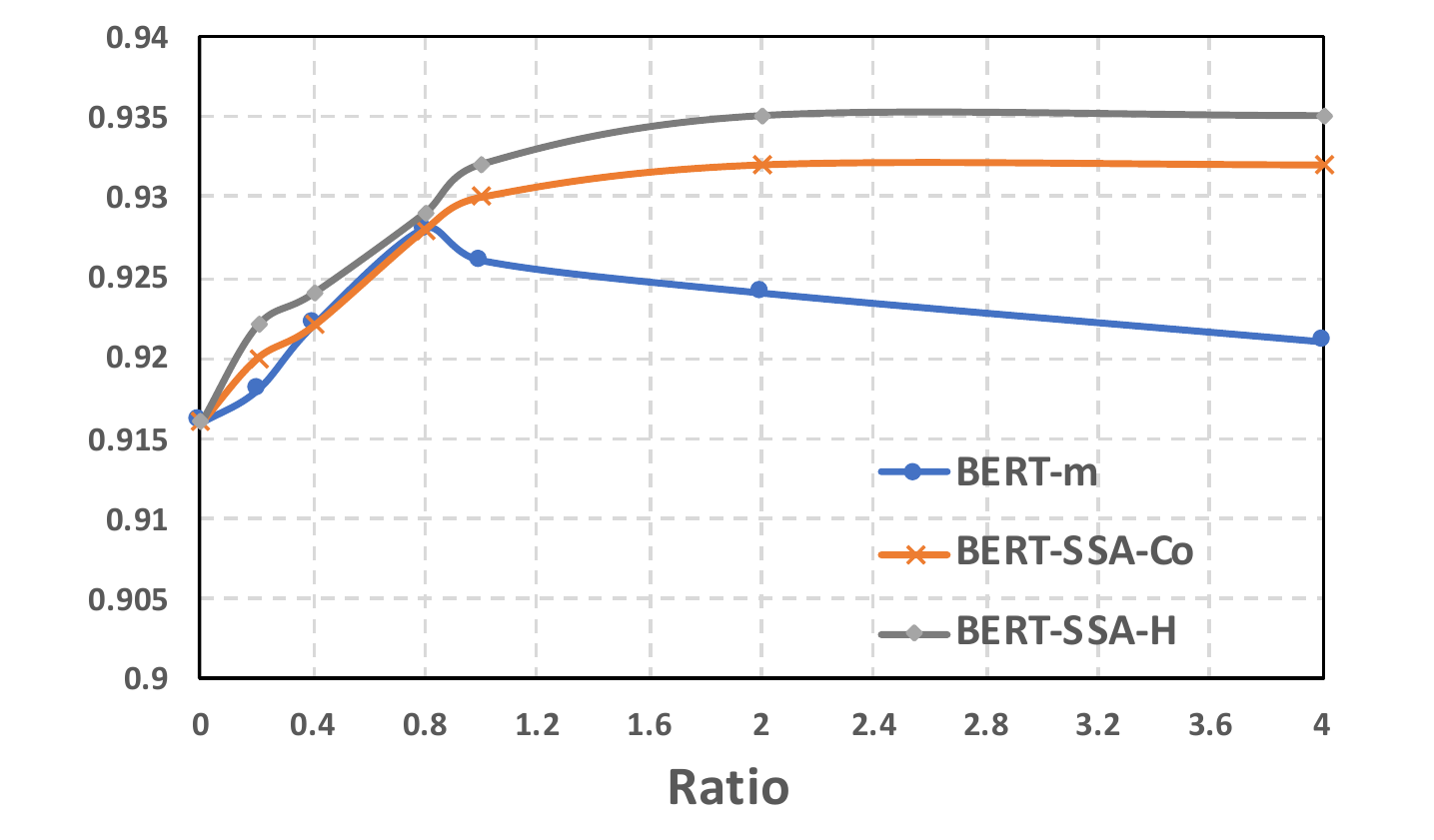}
	\caption{
	Results of parameter sensitivity on SST-2. The x-axis is the value of $\gamma$ and the y-axis is the corresponding accuracy.}
	\label{fig:parameter}
\end{figure}

To investigate the effectiveness of SSA label generation strategy, we conduct experiments on SST-2 under different generation ratio $\gamma$. The larger $\gamma$ means more sentences are generated as training samples. We depict the comparison results in Figure \ref{fig:parameter}. In the beginning, with more sentences generated, all three models continue to improve. After $\gamma$ is greater than 0.8, the performance of BERT-mask degrades dramatically. In contrast, our solution \textbf{BERT-SSA-Co} and \textbf{BERT-SSA-Hybrid} retain the enhancement and converge to better performance. The robustness of our model owes to the self-supervised attention layer which identifies relevant words, and a strict label generation strategy that leverages pre-trained knowledge to obtain self-supervised labels.

\subsection{Case Study}
\begin{figure}[ht]
	\centering
	\subfigure[Case 1]{
        \begin{minipage}[t]{1.0\linewidth}
        \centering
        \includegraphics[width=0.8\linewidth]{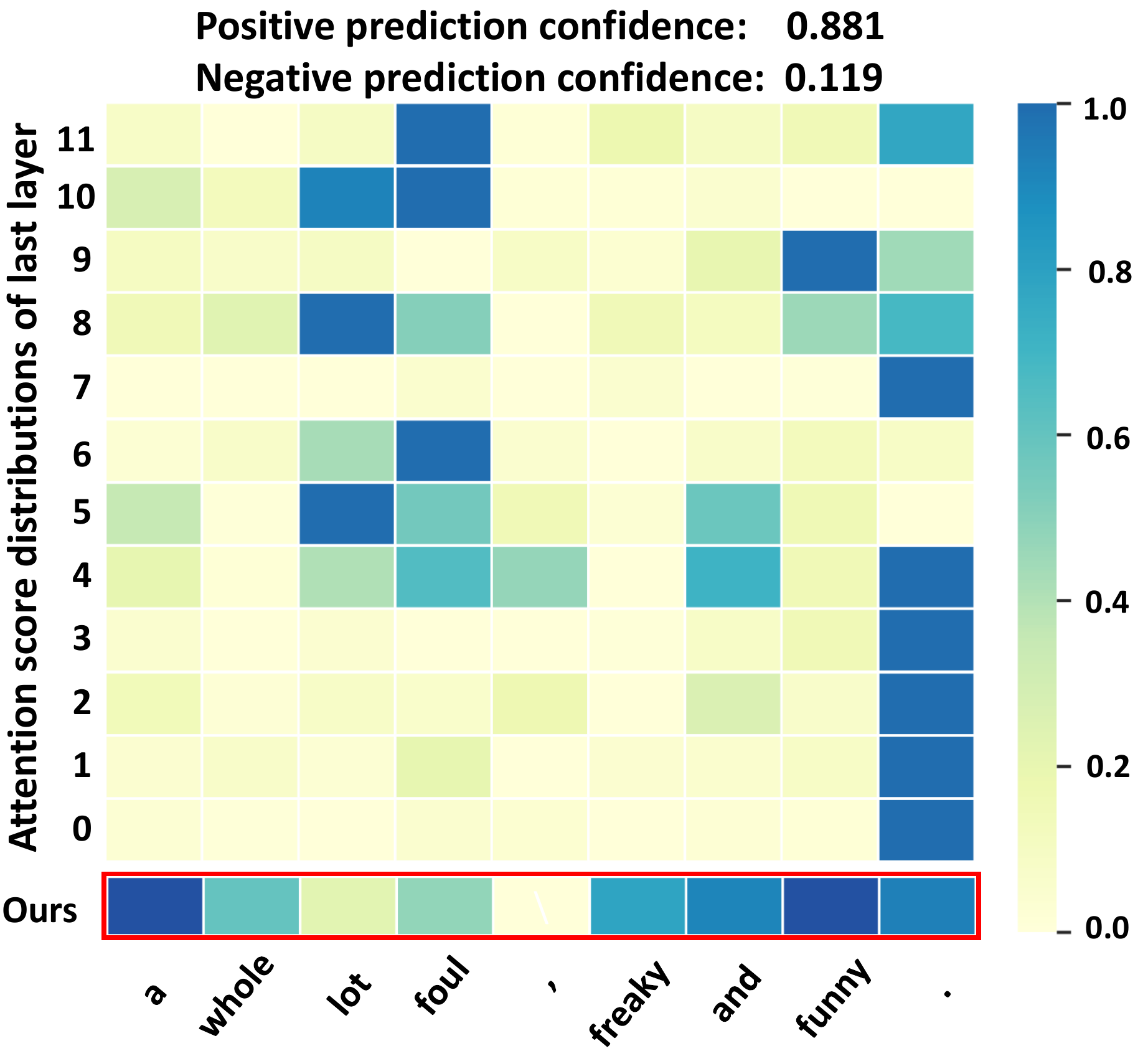}
        \label{fig:case_1}
    \end{minipage}%
    }%
    
    \subfigure[Case 2]{
        \begin{minipage}[t]{1.0\linewidth}
        \centering
        \includegraphics[width=0.8\linewidth]{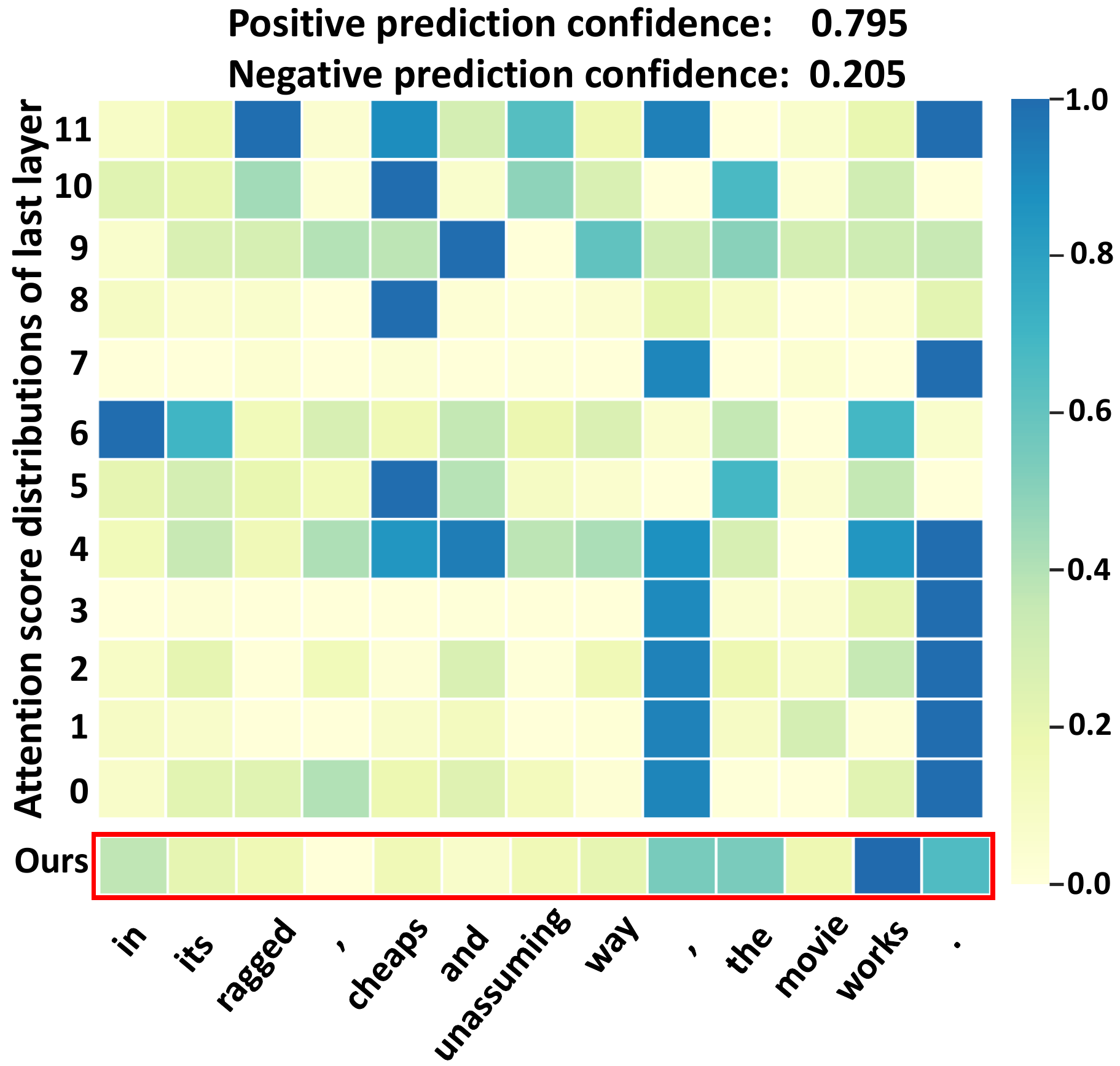}
        \label{fig:case_2}
    \end{minipage}%
    }%
	\caption{
	The multi-head attention scores and SSA score of each word on last layer, obtained by our model fine-tuned on SST.}
	\label{fig:case}
\end{figure}
In this section, we visualize two cases to explain why the self-supervised attention layer improves the model. As illustrated, the SSA scores learned in the hybrid model are in line with human common-sense. One case is the motivating example, i.e., ``a whole lot foul, freaky and funny''. As shown in Figure \ref{fig:case_1}, the vanilla BERT is misled by the strong negative word `foul' and results in a wrong prediction. Instead, Our proposed SSA layer identifies  `funny' as the important token and put less emphasis on the token `foul'. The SSA layer correctly captures the relative word importance and generate better sentence embedding for label prediction. As a result, the sentence is correctly classified as positive by our SSA-enhanced hybrid model. 

Another case is the sentence ``in its ragged, cheaps and unassuming way, the movie works.''. 
It is much more complicated than the first case because more negative words in this sentence may result in the wrong sentiment prediction.
As shown in Figure \ref{fig:case_2}, the vanilla BERT pays much attentions on `ragged' and `cheaps' than `works'. The reason is obvious since `ragged' and `cheaps' present strong negative sentiments while `works' is even not an emotional adjective. However, the self-supervised attention layer is able to identify the actual important tokens, i.e., `works', in such a misleading context. In this way, the final prediction is corrected by the SSA-based model.
\begin{table*}[t]
    \small
    \renewcommand
    \arraystretch{1.2}
    \setlength{\abovecaptionskip}{5pt}
    \begin{tabular}{ccccc}
    \hline
     Index & \multicolumn{1}{m{8cm}}{Sentence} & Target Label & BERT Prediction & SSA-H Prediction \\ 
     \hline
     1 & \multicolumn{1}{m{8cm}}{When your leading ladies are a couple of screen-eating \textcolor{red}{\textbf{dominatrixes}} like Goldie Hawn and Susan Sarandon at their \textcolor{blue}{\textbf{raunchy}} best, even hokum goes down easily.} & Positive & Negative(0.750) & Negative(0.565) \\
     \hline
     2 & \multicolumn{1}{m{8cm}}{Acting , particularly by Tambor , almost makes ``never again’’ \textcolor{blue}{\textbf{worthwhile}}, but (writer/director) Schaeffer should \textcolor{red}{\textbf{follow}} his titular advice.} & Negative & Positive(0.892) & Positive(0.581) \\
     \hline
    3 & \multicolumn{1}{m{8cm}}{I wish I could say ``Thank god, it 's Friday'' , but the \textcolor{red}{\textbf{truth}} of the matter is I was \textcolor{blue}{\textbf{glad}} when it was over.} & Negative & Positive(0.832) & Positive(0.605) \\
     \hline
    \end{tabular}
    \caption{Three examples predicted by original BERT and our SSA-H model on SST-2. The number in parentheses after the label indicates the confidence level. The tokens with the largest attention scores assigned by BERT and SSA-H model are marked with blue and red color respectively.}
    \label{tab:error_analysis}
\end{table*}

\subsection{Error Analysis}
To observe our results in more details, we perform an error analysis for test predictions and show two typical kinds of mistakes below. Table \ref{tab:error_analysis} presents three examples collected from the SST-2 dataset as specific illustrations.

\textbf{The SSA mechanism has weaknesses in commonsense reasoning.} When there are no explicit words relevant to the real label in the sentence, it is difficult to get promotion compared with baseline models. For example, in the first sentence, the SSA-H model assigns a higher attention score to the token `dominatrixes' while BERT focuses on `raunchy'. These two words present negative sentiments and result in a wrong prediction, even though this sentence expresses that actors are good at acting. Similarly, BERT identifies the label of the second sentence as strong positive emotion because of the token `worthwhile'. SSA-H model also makes a mistake in this case. Although our method puts more attention on the word `follow', it could not provide an apparent emotional tendency to get the correct result. Therefore, SSA cannot offer extra background knowledge for commonsense reasoning because all the information comes from the model and the dataset. Under these circumstances, we conjecture a pre-trained model integrating with  domain knowledge will be a better solution.

\textbf{The SSA mechanism obtains limited improvement in capturing intention revealed by consecutive snippets.} Taking the third sentence as an illustration, we notice that BERT and SSA both make wrong predictions. Specifically, BERT assigns the largest attention score on the token 'glad', while SSA takes a step forward by highlighting the word 'truth'. However, for this case, the sentiment revealed by an individual word is not enough to help the model identify the sentimental polarity of the entire sentence. Instead, the intent contained in the snippet ``I was glad when it was over'' is much more critical. Since we use a random mask mechanism, snippets that consist of multiple consecutive words are hardly sampled. Therefore, our future work will focus on improving the efficiency of SSA by designing better sampling strategies to cover various patterns.

\section{Conclusions and Future Work}
In this paper, we propose a novel technique called self-supervised attention (SSA) to prevent BERT from overfitting when fine-tuned on small datasets. The hybrid model contains an additional self-supervised attention layer on top of the vanilla BERT model, which can be trained jointly by an auxiliary SSA task without introducing any external knowledge. We conduct extensive experiments on a variety of neural language understanding tasks, and the results demonstrate the effectiveness of the proposed methodology. The case study further shows that the token-level SSA scores learned in the model are in line with human common-sense. In parallel with our research, many recent works focus on data augmentation and multi-task learning. As the next step, we plan to integrate our methodology with these advanced techniques to achieve the further improvement and evaluate the generality of our proposed solution.

\bibliographystyle{IEEEtran}
\bibliography{access}{}

\begin{IEEEbiography}[{\includegraphics[width=1in,height=1.25in,clip,keepaspectratio]{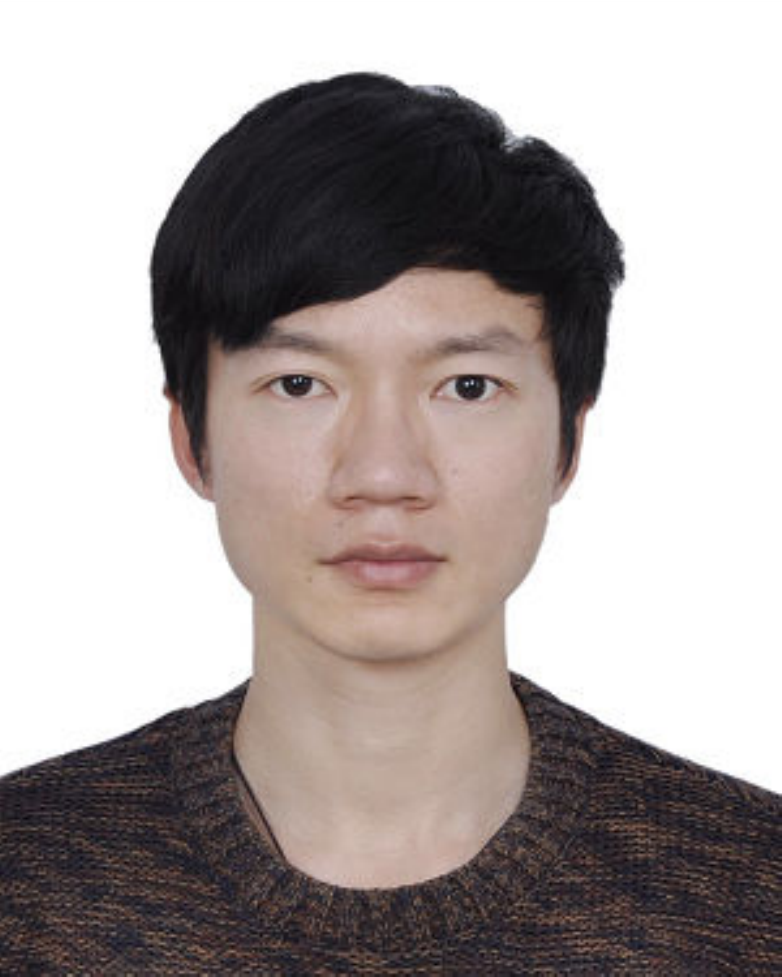}}]{Yiren Chen} received the bachelor's degree from School of Automation Science and Electrical Engineering, Beihang University. He is currently pursuing the Ph.D. degree with School of Electronics Engineering and Computer Science, Peking University. His main research interests include natural language processing and deep learning.
\end{IEEEbiography}

\begin{IEEEbiography}[{\includegraphics[width=1in,height=1.25in,clip,keepaspectratio]{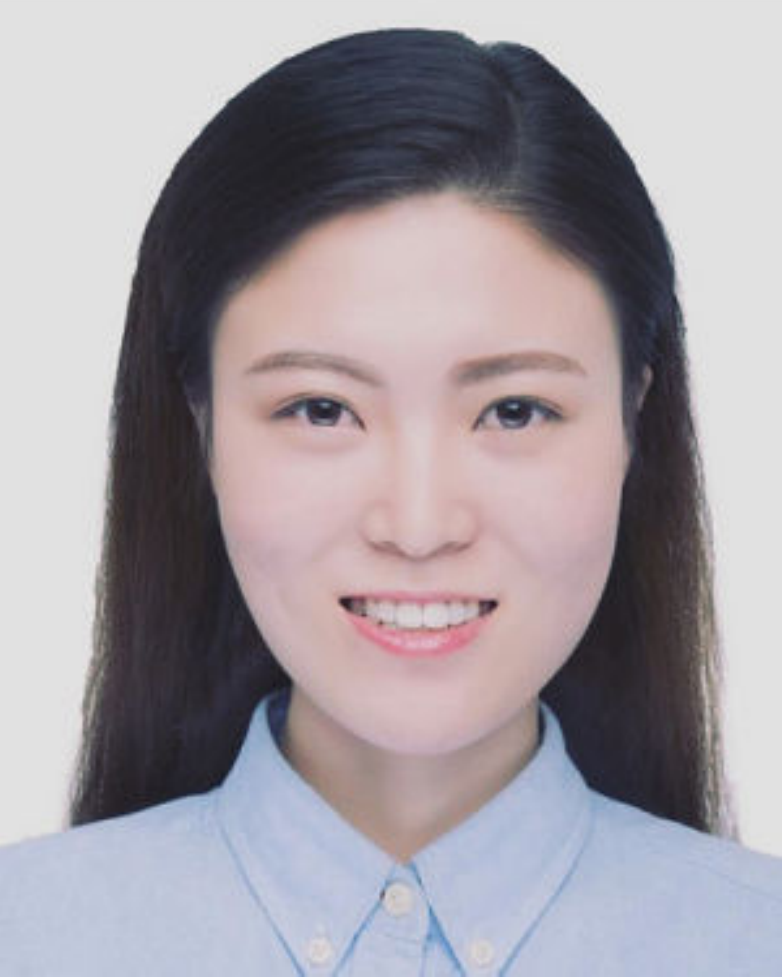}}]{Xiaoyu Kou} received the bachelor's degree from School of Computer Science and Technology, Shandong University and the master's degree from School of Electronics Engineering and Computer Science, Peking University. Her main research interests include knowledge graph and media intelligent computing.
\end{IEEEbiography}

\begin{IEEEbiography}[{\includegraphics[width=1in,height=1.25in,clip,keepaspectratio]{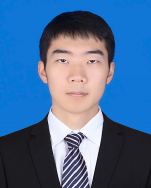}}]{Jiangang Bai} received the bachelor's degree from School of Computer Science and Technology, Sichuan University. He is currently pursuing the Ph.D. degree with School of Electronics Engineering and Computer Science, Peking University. His main research interests include natural language processing and data mining.
\end{IEEEbiography}

\begin{IEEEbiography}[{\includegraphics[width=1in,height=1.25in,clip,keepaspectratio]{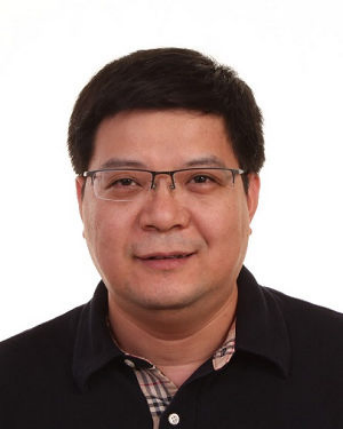}}]{Yunhai Tong} received the Ph.D. degree in computer science from Peking University in 2002. Currently he is a professor in School of Electronics Engineering and Computer Science, Peking University. His main research interests include data mining, media intelligent computing and big data analysis. 
\end{IEEEbiography}

\EOD

\end{document}